\documentclass[runningheads]{llncs}

\usepackage{hyperref}
\usepackage{graphicx}
\usepackage{multicol}
\usepackage{comment}
\usepackage{pdfpages}
\usepackage{todonotes}
\usepackage{xcolor}
\usepackage{amssymb}
\usepackage{multirow}
\usepackage{tabulary}
\usepackage{tikz}
\usetikzlibrary{fit}
\usepackage{algorithm}
\usepackage{algpseudocode}
\usepackage{academicons}

\graphicspath{ {./images/} }

\usepackage{amsmath}
\DeclareMathOperator*{\argmax}{arg\,max}
\DeclareMathOperator*{\argmin}{arg\,min}

\newcommand{\branch}[2]{$b_#1 \in m(a_#2)$}

\begin{document}
\title{Uncertain Machine Ethical Decisions Using Hypothetical Retrospection}

\author{Simon Kolker \orcidID{0000-0001-9144-2856}
\and Louise Dennis\orcidID{0000-0003-1426-1896} 
\and Ramon Fraga Pereira\orcidID{0000-0002-3600-3348}
\and Mengwei Xu\orcidID{0000-0003-4978-3061}}
%
%

\institute{Department of Computer Science, University of Manchester, UK \\
\email{  \{simon.kolker, louise.dennis, ramon.fragapereira, mengwei.xu\}@manchester.ac.uk}
}

\maketitle              

\begin{abstract}
We propose the use of the hypothetical retrospection argumentation procedure, developed by Sven Ove Hansson to improve existing approaches to machine ethical reasoning by accounting for probability and uncertainty from a position of Philosophy that resonates with humans.

Actions are represented with a branching set of potential outcomes, each with a state, utility, and either a numeric or poetic probability estimate.  Actions are chosen based on comparisons between sets of arguments favouring actions from the perspective of their branches, even those branches that led to an undesirable outcome.  This use of arguments allows a variety of philosophical theories for ethical reasoning to be used, potentially in flexible combination with each other.

We implement the procedure, applying consequentialist and deontological ethical theories, independently and concurrently, to an autonomous library system use case.
We introduce a preliminary framework that seems to meet the varied requirements of a machine ethics system: versatility under multiple theories and a resonance with humans that enables transparency and explainability.


\end{abstract}

\section{Introduction}
\label{sec:introduction}
Autonomous machines are an increasingly prevalent feature of the modern world.
From spam filters~\cite{8907831} and fraud detectors~\cite{alhaddad2018artificial}, to drivers~\cite{9016391}, medical practitioners~\cite{yu2018artificial} and soldiers~\cite{szabadfoldi2021artificial}, machines are being developed to automate tasks.
Any decision affecting real people has the potential for ethical impact. Therefore machines are increasingly recognised as ethical agents.
Moor~\cite{moor_2011} categorises such agents as either \emph{implicitly} or \emph{explicitly ethical}.  Implicit ethical agents are built and situated by humans to have a neutral or positive effect, like an ATM machine; they do not utilise concepts of right and wrong in their internal decision making. As autonomous systems make more decisions with more responsibility, they need to reason about ethics \textit{explicitly}. Allen et al. identify two strategies for designing explicitly ethical systems~\cite{allen2005artificial}: \textit{bottom-up} approaches train systems to make ethical decisions with learning techniques based on data from human decision making;
\textit{top-down} approaches encode principles and theories of moral behaviour (often drawn from philosophy) into rules for a selection algorithm, generally using techniques from the field of symbolic Artificial Intelligence (AI).
In this paper, we propose and implement a top-down, explicitly ethical approach.

When an action is taken in the real world, its exact results are typically uncertain. 
As such, a top-down machine ethics system needs a mechanism for handling uncertainty over outcomes.
There are mechanisms for handling uncertainty in AI, including Bayesian methods, Dempster-Shafer theory, fuzzy logics and others~\cite{saffiotti1987ai}.
Nevertheless, it is currently unclear how they might integrate with machine ethics; there may be unanticipated philosophical implications. 

Instead, we opted to operationalise and implement Sven-Ove Hansson's hypothetical retrospection procedure~\cite{hansson2013ethics}.  Originating in Philosophy, the procedure was designed to guide ethical reasoning under uncertainty. 
It favours no specific ethical theory, but systematises the foresight argument pattern, extending an assessor's perspective to judge decisions by the circumstances in which they were made. 
Therefore, arguments can be grounded in a variety of ethical theories.
Over the past ten years, the field of machine ethics has implemented many such theories~\cite{tolmeijer2020implementations}, yet there is no consensus over which is most effective.
Philosophy too has not agreed which is morally correct, leaving implementers to choose from the perspective of stakeholder requirements and preferences. Thus, a mechanism for handling uncertainty that adapts to different ethical theories is desirable.

We outline the procedure via an example from Hansson~\cite{hansson2013ethics}. Suppose an agent is given the choice between an apple and flipping a coin. If the coin lands heads, they win a free holiday to Hawaii. If the coin lands tails, they get nothing.
Selecting the coin is clearly a valid choice.  How might this decision be justified? Under hypothetical retrospection, we list each possible outcome: choosing the apple; choosing to toss the coin and winning the Hawaii holiday; choosing to toss the coin and losing. Next, we \textit{hypothetically retrospect} from each outcome's endpoint. 
Intuitively, the objective is to find an action whose outcomes do not lead the agent to \textit{regret} the ethical implications of their action.\footnote{We recognise there is little ethical impact in this decision, besides maximising utility. It serves as an abstract example where one decision openly defeats another.}
First, consider the coin's outcomes: after winning the holiday, there cannot be regret since the Hawaii holiday is the best outcome; after losing the coin flip, the agent has nothing which is the worst outcome, but there is no regret since the agent justifies that they had a good chance of winning Hawaii, which is far better than an apple.
Now, consider choosing the apple. Here, the agent regrets that they missed a chance of a holiday worth far more than an apple. We saw that choosing the coin did not lead to such regret. Therefore, the procedure advises we pick the coin, matching our intuition.

This paper operationalises the hypothetical retrospection procedure, and the foresight argument pattern it is based on. We implement and evaluate it with moral theories from Philosophy.
We consider Deontology, which specifies a set of actions that are strictly forbidden~\cite{sep-ethics-deontological}, and a theory of consequentialism, which specifies an action is good if its consequences maximise good for the greatest number of people~\cite{Mosdell2011}.
We illustrate our approach with the novel scenario of an autonomous library system. We demonstrate the system's potential for explainability and versatility, while discussing issues and future work.

In Section \ref{sec:RelatedWork}, we will cover related work in the area and highlight this paper's contribution. In Section \ref{sec:background}, we will cover background on symbolic argumentation and uncertainty in Ethical Philosophy. In Section \ref{sec:hypotheticalRetrospection} we will recap Hansson's description of hypothetical retrospection; in Section \ref{implementation} we overview the implementation, including notation, the representation of probability and the argumentation model. Section \ref{sec:testCase} describes our test case of the autonomous library system, its formalism, and our results. Finally, in Section \ref{discussion} we will identify the system's potential benefits and its shortfalls left for future work.

\section{Related Work}
\label{sec:RelatedWork}
This is not the first attempt at building a top-down explicitly ethical machine. Tolmeijer et al. presents an exhaustive survey of implementations as of 2020, but finds the effect of uncertainty is rarely addressed~\cite{tolmeijer2020implementations}. 
Dennis et al. developed a framework suggesting how an autonomous system should act in unforeseen circumstances, with no positive outcomes. However, it does not address uncertainty between the likelihood of outcomes~\cite{dennis2016formal}.
Probabilistic reasoning, such as Bayesian networks~\cite{stephenson2000introduction} and Markov models~\cite{davis2018markov}, has been applied to machine ethics, mostly with regards to maximising expected utility~\cite{cloosUtilibot}. There are a number of criticisms of this approach which we will touch on in Section~\ref{sec:background}.
Killough et al. goes further, architecting agents sensitive to utility risk and reward, with an ability to dynamically adjust risk-tolerance for the environment~\cite{killough2016risk}.

This paper is interested in a framework that incorporates a variety of philosophical ethical theories and allows for the combination of multiple theories, such as Deontology~\cite{sep-ethics-deontological}, Contractualism~\cite{sep-contractualism} and Virtue Ethics~\cite{sep-ethics-virtue}.
Different philosophical theories can advise on different courses of action,  not only in tricky dilemma situations but sometimes even in situations where the moral choice seems intuitively obvious.  There has been some work within machine ethics on comparing and combining different theories. For instance, Sholla et al. weights different principles and then uses fuzzy logic to decide between their recommendations~\cite{sholla2021fuzzy}. Ecoffet and Lehman~\cite{pmlr-v139-ecoffet21a} use a voting procedure in which different ethical theories vote on recommendations but they struggle with the difficulty of comparing Utilitarian theories that return a score for actions with deontological theories that tend to return a judgement that the action is either permissible or impermissible. Our framework enables a flexible approach in which the construction of an argument can treat all ethical theories equally, or allow one to have precedence over another.  The HERA project~\cite{lindner2017hera} is of interest here -- while it does not combine ethical theories it provides a single framework in which many theories can be formalised and operationalised, allowing their recommendations to be compared.  Cointe et. al~\cite{10.5555/2936924.2937086} do something similar with an Answer-Set Programming approach though focused, in this case, on enabling the agent to make moral judgements about others.  These systems could, potentially, be integrated into our argumentation framework to supply judgements on the rightness of an action and its consequences from the perspective of a particular moral theory.

Atkinson and Bench-Capon have developed a framework for ethical argumentation~\cite{Atkinson2016-ATKSGA}. Like our work, assessments of action's outcomes are modelled as arguments.  However, Atkinson and Bench-Capon's work remains concerned with epistemic conflicts between arguments (i.e. disputes between the truth of argument's circumstances) and annotates attacks and defends within the argumentation framework with values, aligning it with the philosophical theory of Virtue Ethics. Our work pivots away, focused purely on the ethical conflicts between arguments. We can assume epistemic truth because arguments are based only on potential, purely hypothetical, versions of events, each created from a single, shared set of information. This allows us to address moral conflict directly. It also lets us build uncertainty into the argumentation mechanism, instead of delegating it to a detail of argument attacks.

\section{Background}
\label{sec:background}
The effect of uncertainty on machine ethics has been relatively unexplored largely due to the lack of research on how uncertainty impacts ethics in general. As Altham explains, there seems to be a gap in moral theory for uncertain situations~\cite{altham1983ethics}. He postulates this could be due to a belief among philosophers that no special principles are required; Moral Philosophy decides the virtues and it is up to Decision Theory to decide how they should be maximised under uncertainty.

 Hansson shows that Utilitarian theories are straightforward in this regard~\cite{hansson2013ethics}. These theories judge decisions based on numeric utilities assigned to their consequences. Expected utility Utilitarianism uses probabilities as weights to discount the utility of improbable outcomes. Hansson critiques this adaptation for the same reason as actual Utilitarianism: its assumption that outcomes can be appraised in terms of a single number (or at least done so both easily and accurately) often produces unintuitive outcomes. In the Apple-Coin scenario from Section \ref{sec:introduction}, although it is evident that a trip to Hawaii holds more value than an apple, the extent of the difference in value remains uncertain. Adding more apples, such as 100, 1000, or 1001, does not necessarily make the deal any more appealing. In other words, apples and holidays are not proportionally comparable. There is no method of assigning relative utilities to all possible states. Brundage briefly surveys other critiques against consequentialist theories. First, they fail to account for personal social commitments, i.e. to friends and family. Second, they do not consider individual differences and rights, tending to favour the majority over any minority. Lastly, they place excessive demands on individuals to contribute to others~\cite{brundage2014limitations}.

Traditional Deontological systems~\cite{sep-ethics-deontological} are made of principles which should never be violated. Hansson shows that any form of probabilistic absolutism, where an action is not permitted if there is any chance of a rule violation, would be too restrictive. Therefore, an approach involving probability thresholds is often suggested. Here, an action is only forbidden when the probability that it violates a law exceeds some limit. The exact value of this limit is open for debate. It is tempting to suggest the limit should have some relation to the action's potential benefits, but this could soon reduce to some elaborate form of Utilitarianism, adamantly against the essence of the original theory.

Noticeably, most humans do not consciously rely on one philosophical, moral theory to make their decisions~\cite{białek_neys_2017}. Nor do we think it is our place to choose a single theory to apply to machine ethics. 
As such, one of Hansson's key contributions is providing an argumentation procedure that can frame multiple, possibly conflicting theories rationally.
To model this, we look to the study of abstract argumentation. Dung creates a framework of logically generated, non-monotonic arguments~\cite{dung1995acceptability}. They can discredit each other with attacks, modelled as a binary relation between the arguments. Dung goes on to specify properties of a well-founded framework; he gives procedures for believing arguments based on their membership to framework extensions. 
This paper will take only take the simple structure of Dung's framework. We leave it to Hansson's philosophy to define attacks and select arguments.

\section{Hypothetical Retrospection}
\label{sec:hypotheticalRetrospection}
Hypothetical retrospection systematises ethical decision making with uncertain outcomes such that its judgements resonate with humans. In this section, we overview Hansson's description of the procedure from \cite{hansson2013ethics}, before we operationalise it in Section \ref{implementation}.

Much of moral philosophy can be interpreted as an attempt to extend a decision maker's perspective. In promoting empathy, we invoke a perspective extending argument pattern to consider other's perceptions of our actions. For cases of uncertainty, Hansson argues it is helpful to extend our perspective with future perceptions of our actions. This means viewing, or hypothetically retrospecting on, a choice from the endpoint of its major foreseeable outcomes. As a result, the hypothetical outcomes, or the \textit{potential branches of future development}, can be used to build resonate arguments about what to do in the present. 
Although Hansson proposes moral arguments that go beyond utility, duty or rights based calculations, the procedure is compatible with many theories of Ethics.

Hansson determines each action's branches of future development like a search problem. Theoretically, a decision's  effects may be infinitely complex and far-reaching. The major search principle, therefore, is to find the most probable future developments which are the most difficult to defend morally. This will increase the chance of considering unethical scenarios. Branches should be developed to an endpoint sufficiently far to capture all morally relevant information. Intermediate information must be captured too: rule violations occurring before the point of retrospection still need to be considered. Additionally, and for the sake of comparison, branches should be described with the same type of information where possible\footnote{The way in which consequences are discussed here may seem to exclude non-consequentialist theories. Hansson emphasizes that this is not the case. In his approach, consequences are broadly defined and their \textit{information} includes agency, virtue intentions, and any other information necessary for moral appraisal.}.
Hansson sees no reason not to create alternate branches based on the uncertainty of the decision maker's own future choices, considering human's inability to control their future actions. Whether an autonomous system has uncertainty over its future actions depends on the nature of the agent and its application architecture.

Our implementation assesses actions assuming their potential branches are provided. In future work, a planning algorithm could be adapted to the requirements above.
 For instance, the Probabilistic Planning Domain Definition Language (PPDDL)~\cite{PPDDL_2004} is able to formalise different stochastic planning settings, e.g., Markov Decision Process (MDP)~\cite{AIJ_HansenZ01a_LAOStar}, Stochastic Shortest Path problems (SSP)~\cite{BertsekasT91_SSP}, and Fully Observable Non-Deterministic planning~\cite{FOND_AIJ_2003_Cimatti}.  This was superseded recently by the Relational Dyanamic Influence Diagram Language (RDDL)~\cite{sanner2010relational} which has been adopted by the International Probabilistic Planning Competition (IPPC)\footnote{\url{https://ataitler.github.io/IPPC2023/}} and is thus the target input language for many planning implementations.

Using their potential branches, actions can be assessed with a selection of ethical theories. Hansson stresses we are not to assess actions in isolation; assessments are purely comparative. This is because decisions are not made in isolation. Given a choice between actions A and B, choosing A is choosing A-instead-of-B. Building action assessments from comparisons ensures all morally relevant information is taken into account.

Actions are compared by hypothetically retrospecting from the endpoint of each action's potential branches of future development. We search for an action which never leads an agent to morally regret its choice in retrospect. Hansson argues against the term \textit{regret} since it is considered a psychological reaction; humans often feel regret for actions they did not commit, or that they could not have known were wrong. By regret, therefore, we mean that the decision making was logically flawed under retrospection. As a result, we use the term \emph{negative retrospection} to reflect this more technical definition. By hypothetically retrospecting between actions' branches, we search for an action which does not lead to negative retrospection, or has full acceptability among its branches. If no such action exists, one should be selected that maximises acceptability in its most probable branches.

Therefore, hypothetical retrospection's decisions are based on relevant ethical information using moral arguments that resonate with humans.

\section{Implementation}
\label{implementation}

\subsection{Formalism}
\label{sec:formalism}
We define an ethical decision problem as a tuple $\langle A, B, S, U, F, I, m\rangle$, composed of an ethical environment and a set of available actions, each with a set of potential branches of future development.

An environment's ethically relevant properties are represented by the set $S$ of Boolean variables; the set $I$ defines the initial truth assignment to $S$, before actions are taken.
For example, in the Coin-Apple scenario there are three state variables in $S$: $s_1$ represents whether or not we have an apple,  $s_2$ whether or not we have gambled, and $s_3$ is whether or not we won a trip to Hawaii.  In the initial state $I$, all these variables are false.

Ethical information for consequentialist and deontological theories are formalised with sets $U$ and $F$. To capture the issue from Section 1, where different event outcomes have an immeasurably greater/lower utility, we have introduced the notion of utility classes.
\begin{definition} [Utility Class] A utility class is an unordered set of individual utility assignments represented as tuples of $\langle s_k, \phi, v \rangle$, where $s_k$ denotes a state variable in $S$ and $v \in \mathbb{R}$ represents the variable's utility when assigned Boolean value $\phi$.
\end{definition}
The ordered set $U$ contains utility classes in descending order of importance. Where $i<j$, all the positive utilities in $u_i$ are considered greater than any utility in $u_{j}$; all the negative utilities in $u_i$ are considered less than any utility in $u_{j}$. To reiterate, the absolute utilities in lower indexed classes are considered immeasurably greater.
In the Coin-Apple example, there are two utility classes in $U$.
The first contains the utility assignment, $\langle s_3, True, 1 \rangle$ representing a utility of 1 for getting the Hawaii holiday.
The second class has utilities immeasurably lower. It contains one assignment, $\langle s_1, True, 1 \rangle$ representing a utility of 1 for getting the apple.

The set $F$ describes the states forbidden by a given deontological theory. This is not the same as defining a negative utility in $U$ since utilities can be outweighed by a greater positive utility. In deterministic decision making environments, forbidden states can not be outweighed. They could represent, for instance, that someone was deceived, that a law (e.g., trespass) was broken, and so on -- any action or outcome that can not be justified. The formalism assumes that the high-level rules have been translated into domain-level rules, applicable to the state variables in $S$. 
\begin{definition}[Forbidden State]
A Forbidden State is a tuple $\langle s, \phi \rangle$ where $s \in S$ is a state variable forbidden from being assigned the Boolean value $\phi$. 
\end{definition}
In the Coin-Apple scenario, $F$ could contain a forbidden state, $\langle s_2, True \rangle$ representing a rule against gambling.

With an environment of ethical values, we define set $A$ of available actions and set $B$ of all potential branches of future development. We define a mapping, $m$, that associates every action with its potential branches of future development. Each branch, $b \in m(a)$ is an ordered sequence of \emph{events} that could occur after action $a$.

\begin{definition} [Event] 
An event is a tuple of $\langle s, \phi, p \rangle$ where $s \in S$, $\phi$ is the new Boolean value of $s$, and $p$ is the probability that the event occurs.

An event therefore represents the change in value of one state variable in $S$.  A branch is a sequence of events that can occur after the action is taken.
\end{definition}

\begin{figure}
    \centering
    \scalebox{1.1}{
        \includegraphics{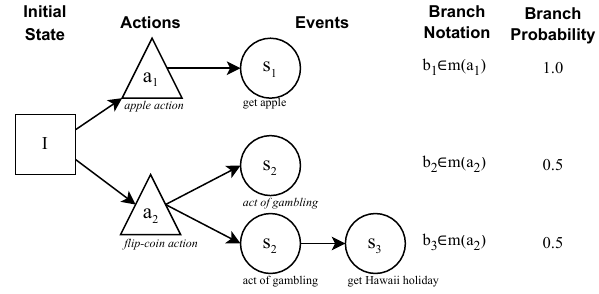}
    }
    \caption{Diagram for Coin-Apple scenario. Event nodes represent True assignment to a state variable. Actions map to a set of branches, represented by rows of event nodes. Probability of conjunction of branchs' events given under branch probability.}
    \label{fig:CoinAppleDiagram}
\end{figure}

For the Coin-Apple example, there are two available actions in~$A$. 
Action $a_1$ represents choosing the apple.  It maps to one branch~\branch{1}{1}, containing one event, $\langle s_1, True, 1 \rangle$---if we choose to have an apple, we gain an apple; we have not gambled nor won a holiday to Hawaii.
Action $a_2$ represents flipping the coin. It maps to two branches, $b_2, b_3 \in m(a_2)$.
The branch $b_2$ contains one event, $\langle s_2, True, 1 \rangle$---we gambled, but we have no apple and no holiday to Hawaii.
The branch $b_3$ is the sequence of events $\langle s_2, True, 1 \rangle$ then $\langle s_3, True, 0.5 \rangle$---first we gambled, then we won a holiday to Hawaii.  The Coin-Apple problem is shown in Figure \ref{fig:CoinAppleDiagram}.

We define the ethical decision problem and a permissible action. The definition of acceptability depends on the ethical theories under consideration (see Section~\ref{sec:argumentation}).

\begin{definition}[Ethical Decision Problem]  An ethical decision problem is a tuple of $\langle A, B, S, U, F, I, m\rangle$ where $A$ stands for a set of available actions, $B$ the set of all potential branches of future development, $S$ the set of Boolean state variables, $U$ an ordered set of \emph{utility classes}, $F$ a set of forbidden state assignments, $I$ the initial assignment of Boolean values to the variables in $S$, representing the initial state, and $m: A \rightarrow {\cal P}(B)$ (where ${\cal P}$ is the powerset function) is a mapping of actions to potential branches of development.
\end{definition}

\begin{definition}[Permissible Action]
    Given an ethical decision problem, defined as a tuple of $\langle A, B, S, U, F, I, m\rangle$, a permissible action is an action, $a \in A$, such that for all potential branches of future development $b \in m(a)$, there is acceptability over their events in state space $S$. If no such actions exist, action $a$ is permissible if it maximises the cumulative probability of its acceptable branches.
\end{definition}

\subsection{Probability Representation}
In many scenarios, while a person may have an intuition that some events are more probable than others, their exact probabilities are unknown. This is most common when interacting with humans and complex systems. Our implementation supports the use of estimative as well as exact probability estimates.
Kent found that intelligence reports tend to use \textit{poetic} words like \textit{probable} or \textit{unlikely}~\cite{kent1964words}. The issue is that people have different interpretations of their meaning. Kent defined a relation for poetic words to mathematical probability ranges, as given in  Table \ref{tab:KentsWords} from~\cite{kent1964words}. Our implementation supports both estimative and exact probabilities.

\begin{table}[]
\begin{center}
\begin{tabular}{|p{0.13\textwidth}|c|c|c|}
 \hline
 \multicolumn{4}{|c|}{100\% Certainty} \\
 \hline
 \multirow{3}{4em}{The General Area of Possibility} & 93\% & Give or take 6\% & Almost Certain\\
 \cline{2-4}
 & 75\% & Give or take 12\% & Probable\\
 \cline{2-4}
 & 50\% & Give or take 10\% & Chances about even\\
 \cline{2-4}
 & 30\% & Give or take 10\% & Probably not\\
 \cline{2-4}
 & 7\% & Give or take 5\% & Almost certainly not\\
 \cline{2-4}
 \hline
 \multicolumn{4}{|c|}{0\% Impossibility}\\
 \hline
\end{tabular}
\caption{Mathematical to poetic relation from Kent's estimative probability~\cite{kent1964words}.}
\label{tab:KentsWords}
\end{center}
\end{table}

\subsection{Argumentation Model}
\label{sec:argumentation}
Hansson does not give steps for comparing action's potential branches of future development in~\cite{hansson2013ethics}. For our implementation, we chose to build comparative moral assessments with a simple argumentation network, based partially on the work of Atkinson et al.~\cite{atkinson2004justifying}.
Here, arguments are generated logically from an \emph{argument scheme}. For an action $a \in A$, selected in initial state $I$, resulting in the branch \branch{{}}{{}} with probability $p$, the following argument is generated:
\begin{center}
     \textit{``From the initial state $I$, it was acceptable to perform action $a$, resulting in consequences $b$ with probability p."}
\end{center}
For notation, this is written $Argument(b)$.  We view this as a default argument that any action is acceptable.
In our running example, the retrospective argument below is generated for $b_3$, tossing the coin and winning the Hawaii holiday.

\begin{center}
    \textit{``From the initial state $I$, where $s_1=s_2=s_3=False$, it was acceptable to perform the action $a_2$, resulting in consequences with $s_2=s_3 = True$ with probability 0.5."}
\end{center}

To determine an argument's validity, we search for attacks from other actions' arguments. Incoming attacks imply negative retrospection for not choosing an attacking action. To formalise Hansson's retrospection, we generate attacks by posing critical questions on arguments' claims~\cite{atkinson2004justifying}. For the branches $b_1 \in m(a_1)$, $b_2 \in m(a_2)$ and any generic moral principle, the following critical questions are asked for $Argument(b_1)$ to attack $Argument(b_2)$.
\begin{description}
    \item[CQ1] \textit{Did $b_2$ violate a moral principle that $b_1$ did not}?
    \item[CQ2] \textit{Did $a_2$ hold a greater probability of breaking the moral principle than $a_1$}?
\end{description}
$Argument(b_1)$ only attacks $Argument(b_2)$ if both of these questions are answered positively. They represent negative retrospection for missing the chance to avoid violating a principle.
The critical questions are asked both ways between all arguments supporting different actions, for every moral principle under consideration. The time and space complexity of answering the questions will differ for different theories.
The desired ethical theories have to be encoded into the critical questions relative to a domain. For Utilitarianism and a generic deontological do-no-harm principle critical questions are embedded as follows:
\begin{itemize}
    \item Utilitarian CQ1: \textit{Did $b_2$ bring greater utility value than $b_1$}?
    \item Utilitarian CQ2: \textit{Did $a_2$ expect greater utility value than $a_1$}?
    \item Do-no-harm CQ1: \textit{Did $b_2$ cause harm where $b_1$ did not}?
    \item Do-no-harm CQ2: \textit{Did $a_2$ expect greater probability of causing harm than $a_1$}?
\end{itemize}
After searching for attacks on all branches, an action should be selected with complete acceptability. If no such action exists, an action should be selected with maximal acceptability, i.e. summing the probability of each non-attacked argument and selecting an action with a maximal sum.

\subsection{Algorithm}
We outline our implementation in Algorithm~\ref{mainAlgo}. Given an ethical decision problem, all actions are compared by their potential branches of future development (lines 2-4). There is a hypothetical retrospective argument made from the perspective of each branch in favour of its action. Attacks are generated between arguments by asking two critical questions based on an ethical theory.  For our implementation we use a Utilitarian and a Deontological theory (lines 5-6), detailed later in Algorithm~\ref{utilAlgo} and \ref{deonAlgo}. Attacked branches are marked as such (lines 7-13). An action's acceptability defaults to 1 and is subtracted by the cumulative probability of attacked branches. The action with maximum acceptability is selected (lines 17-25).

\begin{algorithm}[t]
    \caption{Arguments action's potential branches of future development. Returns index of action with maximum acceptability.}\label{alg:cap}
    \textbf{Input} Ethical Decision Problem $\langle A, B, S, U, F, I, m\rangle$\\
\textbf{Output} Permissible Action $a \in A$

\begin{algorithmic}[1]
\State \textbf{array} $attacked$ $\gets$ $[False,...,False]$ \textbf{of size} $length(B)$
\For{$\textbf{each}$ $a_i, a_j$ $\textbf{in}$ \{$(a_i,a_j) | a_i,a_j\in A$ $\textbf{and}$ $a_i \neq a_j$\}}

\For{$\textbf{each}$ $b_k$ $\textbf{in}$ $m(a_i)$}
    \For{$\textbf{each}$ $b_l$ $\textbf{in}$ $m(a_j)$}
        \State uTarget $\gets$ Target in Utilitarian CQs ($b_k \in m(a_i)$, $b_l \in m(a_j)$, $U$)
        \State dTarget $\gets$ Target in Deontological CQs ($b_k \in m(a_i)$, $b_l \in m(a_j)$, $F$)
    
        \If{dTarget $==$ uTarget \textbf{and} dTarget \textbf{is} not None}
            \State attacked[uTarget] $\gets True$
        \ElsIf{dTarget $!=$ uTarget \textbf{and} dTarget \textbf{is} None}
            \State attacked[uTarget] $\gets True$
        \ElsIf{dTarget $!=$ uTarget \textbf{and} uTarget \textbf{is} None}
            \State attacked[dTarget] $\gets True$ 
        \EndIf
    \EndFor
\EndFor
\EndFor
\State \textbf{array} $acceptability$ $\gets$ $[1,...,1]$ \textbf{of size} $length(A)$
\For{$\textbf{each}$ $a_i \in A$}
    \For{$\textbf{each}$ $b_k \in m(a_i)$}
        \If{$attacked[k]$}
            \State $acceptability[i]$ $\gets$ $acceptability[i] - Probability(b_k)$
        \EndIf
    \EndFor
\EndFor

\State \textbf{return} $\gets$ $\argmax_i(acceptability[i])$
\end{algorithmic}

    \label{mainAlgo}
\end{algorithm}

Algorithm~\ref{utilAlgo} embeds the theory of Utilitarianism into the critical questions. As explained in Section~\ref{sec:formalism}, branches are made from a list of events which each change a Boolean state variable with some probability. Variable utilities are defined by a set of utility classes, with assignments in lower indexed classes immeasurably greater. Algorithm~\ref{utilAlgo} compares two potential branches and returns the index of a branch if it is defeated by the other branch through the critical questions. It is invoked by Algorithm~\ref{mainAlgo} on line 5.
Algorithm~\ref{utilAlgo} counts from the lowest utility class upwards to find the first class where branch utilities are unequal. If found, critical question 1 is answered positively. The branch with the greater utility becomes the \textit{attacker}, the other is the \textit{defender} (lines 1-9). If utilities are equal through all classes, there are no attacks (lines 10-12). Otherwise, the defender branch attempts to use the foresight argument to defend itself: for each lower indexed class, if the defender's action has greater expected utility, defence is successful and there is no attack (lines 13-17). If the attacker action has greater or equal expected utility across all classes, defence fails and critical question 2 is positive. Thus, the defender branch is attacked (line 18).

\begin{algorithm}[]
    \caption{For two potential branches of future development, finds target with lower utility in utility classes and no greater utility expectation to defend.}\label{alg:cap}
    \textbf{Input} Action Branches $b_k \in m(a_i), b_l \in m(a_j)$, Utility Classes $U$\\
\textbf{Output} Index of Attacked Branch $x$

\begin{algorithmic}[1]
\For{$c \gets 0$ to $length(U)$}
    \State $value[i] \gets$ Utility of $b_k$ in $U[c]$
    \State $value[j] \gets$ Utility of $b_l$ in $U[c]$
    \If{$value[i]$ \textbf{is} not $value[j]$}
        \State $attacker \gets \argmax_x(value[x])$
        \State $defender \gets \argmin_x(value[x])$
        \State break
    \EndIf
\EndFor
\If{$attacker$ \textbf{is} None}
    \State \textbf{return} $\gets$ None
\EndIf
\For{lowerc $\gets 0$ to $c$}
    \If{Expected Utility of $a_{attacker}$ in $U[lowerc] <$ Expected Utility of $a_{defender}$ in $U[lowerc]$}
        \State \textbf{return} $\gets$ None
    \EndIf
\EndFor
\State \textbf{return} $\gets$ defender
\end{algorithmic}

    \label{utilAlgo}
\end{algorithm}

Algorithm~\ref{deonAlgo} shows Deontology embedded into the critical questions, similar to Algorithm~\ref{utilAlgo}. Algorithm~\ref{deonAlgo} iterates across the set of forbidden assignments and checks the events in either for a violation (lines 1-3). See Section~\ref{sec:formalism} for forbidden assignments. If one branch has a violation that the other does not, then critical question 1 is positive (line 4 and 9). To defend itself, the violating branch's action must have a greater probability of not making the assignment. If this is not true, critical question 2 is positive and the index of the violating branch is returned (lines 4-13). If no branch is attacked, neither index is returned (line 15).

\begin{algorithm}[]
    \caption{For two potential branches of future development, finds target which breaks a deontological law with no greater expectation otherwise.}\label{alg:cap}
    \textbf{Input} Action Branches $b_k \in m(A_i), b_l \in m(A_j)$, Forbidden States $F$\\
\textbf{Output} Index of Attacked Branch $x$

\begin{algorithmic}[1]
\For{\textbf{each} $\langle s, \phi \rangle$ in $F$}
    \State $violation[i] \gets$ Do events in $b_k$ set $s=\phi$
    \State $violation[j] \gets$ Do events in $b_l$ set $s=\phi$
    \If{$violation[i]$ and not $violation[j]$}
        \If{Probability of $s=\phi$ in $m(a_{j})$ $<$ Probability of $s=\phi$ in $m(a_{i})$}
            \State \textbf{return} $\gets$ i
        \EndIf
    \EndIf
    \If{$violation[j]$ and not $violation[i]$}
        \If{Probability of $s=\phi$ in $m(a_{i})$ $<$ Probability of $s=\phi$ in $m(a_{j})$}
            \State \textbf{return} $\gets$ j
        \EndIf
    \EndIf
\EndFor
\State \textbf{return} $\gets$ None
\end{algorithmic}

    \label{deonAlgo}
\end{algorithm}

Our implementation has no planning element, searching for action's branches as discussed in Section \ref{sec:hypotheticalRetrospection}. This is left for future work.
Instead, we pass an ethical decision problem to an implementation of Algorithm 1 and a permissible action is output.
We implement a web app with Flask and Python 3.8.9 to graph retrospection and alter utilities and deontological laws.  The source code is available on GitHub at \url{https://github.com/sameysimon/HypotheticalRetrospectionMachine}.

\section{Autonomous Library Test Case}
\label{sec:testCase}
To demonstrate our implementation, we present an uncertain ethical decision problem and discuss our implementation's selected action given five sets of ethical considerations.

Suppose a student logs onto their University's autonomous library to revise for a test the next morning. All the other students started revision a month ago. As the student constructs various search terms for a recommendation, the system recognises that all other students have taken out the same book, implying it is very useful.
Should the autonomous library use this data to recommend the book, allowing the student to revise quicker on the night before the test? If other students find out, they may feel unfairly treated; students who wait for a reference would get the same credit as those who find it themselves. 

\begin{figure}
    \centering
    \scalebox{1.2}{
\includegraphics{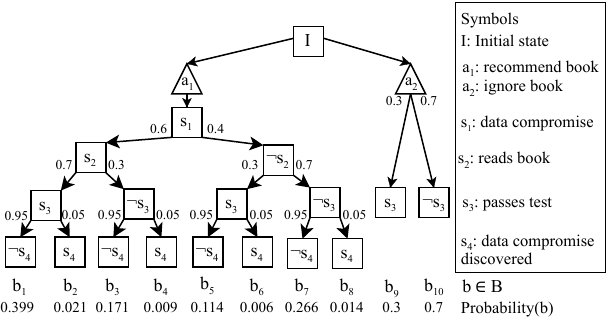}
    }
    
    \caption{Decision tree of possible events in Autonomous Library problem. Triangles represent actions and boxes variable assignments, ¬ represents $False$ assignment.}
    \label{fig:LibraryDiagram}
\end{figure}

We model the scenario as an ethical decision problem, $\langle A,B,S,U,F,I,m \rangle$, with two actions in $A$ mapping to ten branches in $B$, acting across four state variables in $S$.
For action $a_1$, to \textit{recommend} the book, student data is compromised, the truth of which is represented by Boolean variable $s_1$. Given a recommendation, there is a 0.6 chance the book is used, represented by $s_2$. If they have the book, there is a 0.7 chance they will pass, $s_3$, otherwise without the book there is a 0.3 chance they will pass, $s_3$. Finally, there is a 0.05 chance other students will find out their data was compromised, $s_4$. If the system ignores the book, with action $a_2$, there is a 0.3 chance the student will pass, again represented as $s_3$
\footnote{There is discourse on whether a decision to act should be judged the same as a decision not to act~\cite{Foot1967-FOOTPO-2}. We consider ignoring the book an action, an act of discrimination for example, which is assessed the same as the act to recommend.}.
Figure \ref{fig:LibraryDiagram} is a decision tree labelled with probabilities and branch notation.
An argument is generated from each branch's endpoint, representing positive retrospection. Using the argument scheme from Section \ref{sec:argumentation}, $Argument(b_1)$ is the following:
\begin{center}
    \textit{``From the initial state, $I$, where $s_1=s_2=s_3=s_4=False$, it was acceptable to perform the action, $a_1$, resulting in consequences with $s_1=s_2=s_3=True$ and $s_4=False$, with probability 0.399."}
\end{center}

The argument claims it was acceptable to recommend the book, resulting in a data protection violation ($s_1$), the student reading the book ($s_2$) and passing the test ($s_3$), with the data breach kept a secret ($s_4=False$), at a probability of 0.399.

\subsection{Consequentialism with One Assignment}
\begin{figure}[h]
    \usetikzlibrary{fit, shapes, positioning}

\begin{tikzpicture}[]
  \centering
  \node[draw, minimum width=9cm, minimum height=1.3cm, inner sep=1pt, label=left:Recommend book] (box1) {};
  \node[draw, circle] (a1b1) at (-3.7,0.18) {$b_1$};
  \node[draw, circle, right=0.35 of a1b1] (a1b2) {$b_2$};
  \node[draw, circle, right=0.35 of a1b2] (a1b3) {$b_3$};
  \node[draw, circle, right=0.35 of a1b3] (a1b4) {$b_4$};
  \node[draw, circle, right=0.35 of a1b4] (a1b5) {$b_5$};
  \node[draw, circle, right=0.35 of a1b5] (a1b6) {$b_6$};
  \node[draw, circle, right=0.35 of a1b6] (a1b7) {$b_7$};
  \node[draw, circle, right=0.35 of a1b7] (a1b8) {$b_8$};

  \node[draw, minimum width=9cm, minimum height=1.3cm, inner sep=1pt, below=0cm of box1.south, label=left:Ignore book] (box2) {};
  
  \node[draw, circle, below=0.8 of a1b3] (a2b1) {$b_9$};
  \node[draw, circle, below=0.8 of a1b6] (a2b2) {$b_{10}$};
  
  \draw[->, thick, black, -latex] (a1b1.south) -- (a2b2.west);
  \draw[->, thick, black, -latex] (a1b2) -- (a2b2);
  \draw[->, thick, black, -latex] (a1b5) -- (a2b2);
  \draw[->, thick, black, -latex] (a1b6) -- (a2b2);

\end{tikzpicture}
    \centering
    \caption{Graph of retrospection between hypothetical branches of development with only the utility of the student passing in consideration. Incoming edges on an argument represent negative retrospection for not selecting the attacking argument's action.}
    \label{fig:UtilityPassedOne}
\end{figure}
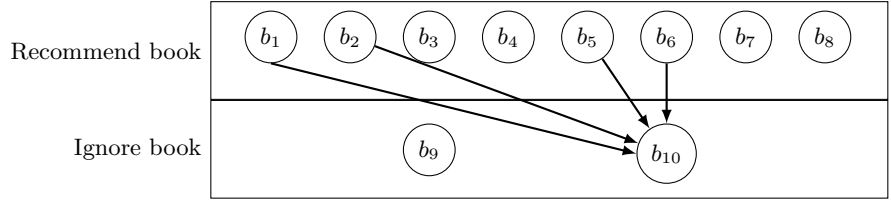

First we test our implementation considering the ethical theory of consequentialism. We set $U$ to have one utility class with one utility assignment, \\$\langle passesTest, 1, True \rangle$. 
The only value is the student passing. Intuitively, the action maximising the probability of passing should be chosen; hypothetical retrospection agrees. The argumentation graph in Figure~\ref{fig:UtilityPassedOne} shows the retrospection.

Every branch has acceptability, except \branch{{10}}{2} where the student fails after the system choo2ses \textit{ignore}, with 0 utility and 0.3 probability (\textit{`probably not'} in Kent's words). This branch has a lower utility than the four \textit{recommend} branches where the student passes: $b_1, b_2, b_5, b_6 \in m(a_1)$. They cause $Argument(b_{10})$ to answer critical question 1 positively when attacked by these arguments.  Since \textit{recommend} has a greater utility expectation, or a greater probability of the student passing, $Argument(b_{10})$ cannot defend itself in critical question 2. Thus, there is no reason to select \textit{ignore}; from the perspective of $b_{10}$'s endpoint there is negative retrospection. There are no other attacks. Therefore by hypothetical retrospection action $a_1$, \textit{recommend}, should be selected.

\subsection{Consequentialism with Two Equal Assignments}
\begin{figure}[h]
    \centering
    \scalebox{0.9}{
        \usetikzlibrary{fit, shapes}

\begin{tikzpicture}[]
  \centering
  \node[draw, minimum width=9cm, minimum height=1.2cm, inner sep=1pt, label=left:Recommend book] (box1) {};
  \node[draw, circle] (a1b1) at (-3.7,0) {$b_1$};
  \node[draw, circle, right=0.35 of a1b1] (a1b2) {$b_2$};
  \node[draw, circle, right=0.35 of a1b2] (a1b3) {$b_3$};
  \node[draw, circle, right=0.35 of a1b3] (a1b4) {$b_4$};
  \node[draw, circle, right=0.35 of a1b4] (a1b5) {$b_5$};
  \node[draw, circle, right=0.35 of a1b5] (a1b6) {$b_6$};
  \node[draw, circle, right=0.35 of a1b6] (a1b7) {$b_7$};
  \node[draw, circle, right=0.35 of a1b7] (a1b8) {$b_8$};

  \node[draw, minimum width=9cm, minimum height=1.2cm, inner sep=1pt, below=0cm of box1.south, label=left:Ignore book] (box2) {};
  
  \node[draw, circle, below=0.5 of a1b3] (a2b1) {$b_9$};
  \node[draw, circle, below=0.5 of a1b6] (a2b2) {$b_{10}$};
  
  \draw[->, thick, black, -latex] (a1b1.south) -- (a2b2.west);
  \draw[->, thick, black, -latex] (a1b5) -- (a2b2);

\end{tikzpicture}
    }
    \caption{Graph of retrospection between hypothetical branches of development with the cost of others finding out data was compromised equaling the utility of the student passing.}
    \label{fig:UtilityPass1Others-1}
\end{figure}
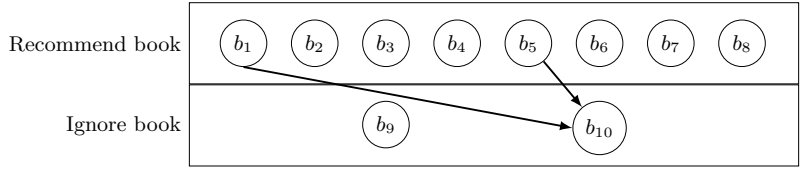

Now we consider two utility assignments of the same class: $\langle passesTest, 1, True \rangle$ and $\langle othersFindOut, -1, True \rangle$.
This invokes the risk of others finding out their data was used, with others finding out judged as bad as the student passing is good. Retrospection is shown in Figure~\ref{fig:UtilityPass1Others-1}.

Again, only branch \branch{{10}}{2} has negative retrospection, when the student fails after the system chooses to \textit{ignore} the book.
This time only two of \textit{recommend}'s branches have greater utility, $b_1, b_5 \in m(a_1)$. Action \textit{recommend} still has a greater utility expectation, so \textit{ignore} cannot be defended in critical question 2. Therefore, \textit{recommend} is selected.

\subsection{Consequentialism with Unequal Assignments}
\begin{figure}[h]
    \centering
    \scalebox{0.9}{
        \usetikzlibrary{fit, shapes}

\begin{tikzpicture}[]
  \centering
  \node[draw, minimum width=9cm, minimum height=1.5cm, inner sep=1pt, label=left:Recommend book] (box1) {};
  \node[draw, circle] (a1b1) at (-3.7,0.3) {$b_1$};
  \node[draw, circle, right=0.35 of a1b1] (a1b2) {$b_2$};
  \node[draw, circle, right=0.35 of a1b2] (a1b3) {$b_3$};
  \node[draw, circle, right=0.35 of a1b3] (a1b4) {$b_4$};
  \node[draw, circle, right=0.35 of a1b4] (a1b5) {$b_5$};
  \node[draw, circle, right=0.35 of a1b5] (a1b6) {$b_6$};
  \node[draw, circle, right=0.35 of a1b6] (a1b7) {$b_7$};
  \node[draw, circle, right=0.35 of a1b7] (a1b8) {$b_8$};

  \node[draw, minimum width=9cm, minimum height=1.5cm, inner sep=1pt, below=0cm of box1.south, label=left:Ignore book] (box2) {};
  
  \node[draw, circle, below=1.3 of a1b3] (a2b1) {$b_9$};
  \node[draw, circle, below=1.3 of a1b6] (a2b2) {$b_{10}$};
  
  \draw[->, thick, black, -latex] (a2b1) -- (a1b2);
  \draw[->, thick, black, -latex] (a2b1) -- (a1b4);
  \draw[->, thick, black, -latex] (a2b1) -- (a1b6);
  \draw[->, thick, black, -latex] (a2b1) -- (a1b8);

  \draw[->, thick, black, -latex] (a2b2) -- (a1b2);
  \draw[->, thick, black, -latex] (a2b2) -- (a1b4);
  \draw[->, thick, black, -latex] (a2b2) -- (a1b6);
  \draw[->, thick, black, -latex] (a2b2) -- (a1b8);

\end{tikzpicture}
    }
    
    \caption{Graph of retrospection between hypothetical branches of development with the cost of others finding data was compromised outweighing the utility of passing.}
    \label{fig:UtilityPass1Others-5}
\end{figure}
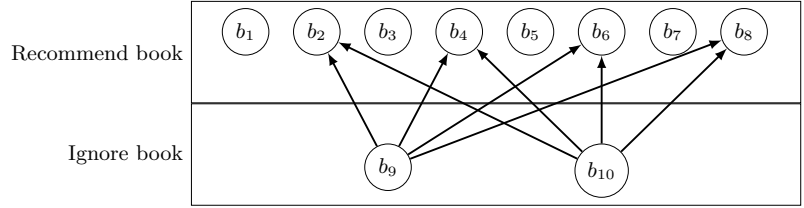

The utility of students discovering the data compromise can be lowered such that \textit{recommend}'s expected utility is lower than \textit{ignore}'s, for example with the assignment $\langle othersFindOut, -5, True \rangle$. 
Now, attacks fire the other way, displayed in Figure \ref{fig:UtilityPass1Others-5}.
When \textit{recommend} is chosen and other students find out, as in $b_2, b_4, b_6, b_8 \in m(a_1)$, the utility is lower than \textit{ignore}'s branches. This answers critical question 1 positively for attacks on these branch's arguments.
There is no defence since \textit{ignore} has a greater utility expectation so critical question 2 is positive. \textit{Recommend} can lead to the highest utility branches with $b_1$ and $b_5$, but unlike before, $b_{10}$ defends citing its higher utility expectation. Thus, \textit{ignore} is selected with full acceptability.

Deciding utilities is difficult without further details, i.e. the student's grades, data preferences, etc. Ideally, branches would be developed until enough morally relevant information is described, but this is not always computationally viable.
Even so, exact utilities are subjective. We confront this issue with utility classes. 
Supposing $othersFindOut$ has utility immeasurably lower than  $passesTest$, we form two classes. The first has assignment $\langle othersFindOut, -1, True \rangle$; the second has $\langle passesTest, 1, True \rangle$. The resulting retrospection is the same as in Figure \ref{fig:UtilityPass1Others-5}, with the cost of others' knowledge outweighing the benefits of passing.

\subsection{Deontology with Consequentialism}

\begin{figure}[h!]
    \centering
    \scalebox{0.9}{
        \usetikzlibrary{fit, shapes, positioning}

\begin{tikzpicture}[
arg/.style={draw, minimum width=0.8cm, circle},
]
  \centering
  \node[draw, minimum width=10cm, minimum height=1.5cm, inner sep=1pt, label=above:Recommend book] (box1) {};
  \node[arg] (a1b1) at (-4.1,0.12) {$b_1$};
  \node[arg, right=0.35 of a1b1] (a1b2) {$b_2$};
  \node[arg, right=0.35 of a1b2] (a1b3) {$b_3$};
  \node[arg, right=0.35 of a1b3] (a1b4) {$b_4$};
  \node[arg, right=0.35 of a1b4] (a1b5) {$b_5$};
  \node[arg, right=0.35 of a1b5] (a1b6) {$b_6$};
  \node[arg, right=0.35 of a1b6] (a1b7) {$b_7$};
  \node[arg, right=0.35 of a1b7] (a1b8) {$b_8$};

  \node[draw, minimum width=10cm, minimum height=1.5cm, inner sep=1pt, below=0cm of box1.south, label=below:Ignore book] (box2) {};
  
  \node[arg, below=1 of a1b3] (a2b1) {$b_9$};
  \node[arg, below=1 of a1b6] (a2b2) {$b_{10}$};
  

  \draw[->, thin, black, -latex] (a2b1) -- (a1b1);
  \draw[->, thin, black, -latex] (a2b1) -- (a1b2);
  \draw[->, thin, black, -latex] (a2b1) -- (a1b3);
  \draw[->, thin, black, -latex] (a2b1) -- (a1b4);
  \draw[->, thin, black, -latex] (a2b1) -- (a1b5);
  \draw[->, thin, black, -latex] (a2b1) -- (a1b6);
  \draw[->, thin, black, -latex] (a2b1) -- (a1b7);
  \draw[->, thin, black, -latex] (a2b1) -- (a1b8);

  \draw[->, thin, black, -latex] (a2b2) -- (a1b1);
  \draw[->, thin, black, -latex] (a2b2) -- (a1b2);
  \draw[->, thin, black, -latex] (a2b2) -- (a1b3);
  \draw[->, thin, black, -latex] (a2b2) -- (a1b4);
  \draw[->, thin, black, -latex] (a2b2) -- (a1b5);
  \draw[->, thin, black, -latex] (a2b2) -- (a1b6);
  \draw[->, thin, black, -latex] (a2b2) -- (a1b7);
  \draw[->, thin, black, -latex] (a2b2) -- (a1b8);

  \draw[->, thick, blue, -latex, dashed] (a1b1) -- (a2b2.west);
  \draw[->, thick, blue, -latex, dashed] (a1b2) -- (a2b2);
  \draw[->, thick, blue, -latex, dashed] (a1b5.south) -- (a2b2);
  \draw[->, thick, blue, -latex, dashed] (a1b6.south) -- (a2b2);
\end{tikzpicture}
    }
    
    \caption{Graph of retrospection between potential branches of development with one Consequentialist assignment and one Deontological law. Consequentialist attacks are dashed blue; Deontological attacks are solid black.}
    \label{fig:DeonLaw}
\end{figure}
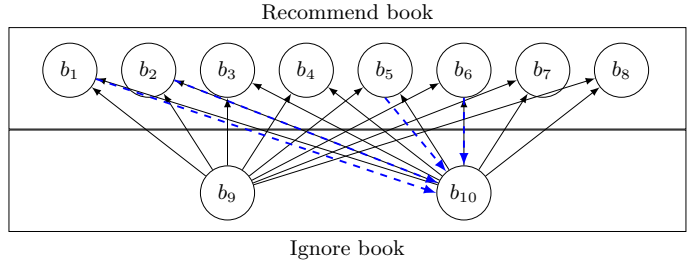

Finally we consider a deontological theory against the misuse of others' data.
This could be the UK Law, requiring under the Data Protection Act that personal data is to only be used for specified, explicit purposes~\cite{dataProtection2018}.
Otherwise, there could be a violation of the Doctrine of Double Effect, having four conditions~\cite{mangan1949historical}: 1. that the action in itself from its very object be good or at least indifferent; 2. that the good effect and not the evil effect be intended; 3. that the good effect be not produced by means of the evil effect; 4. that there be a proportionately grave reason for permitting the evil effect.
If we consider non-consensual use of students' data as bad and helping a student to pass the exam to be good, then the fact that the bad effect is required in order to bring about the good effect breaks the third condition above, and, therefore, is not permissible.
We build on our first test in Section 6.1 which selected \textit{recommend} with utility assignment $\langle passesTest, 1, True \rangle$. Adding forbidden state $\langle dataProtectionViolation, True \rangle$ to $F$ results in the retrospection shown by Figure \ref{fig:DeonLaw}.
Every argument from \textit{ignore} attacks every argument from \textit{recommend} since \textit{ignore} avoids violating the law.

Under our previous Consequentialism, \textit{recommend} is still chosen with the same attacks on $Argument(b_{10})$ as before. This conflict represents a moral dilemma, where no choice is normatively inferior to another~\cite{hansson2013ethics}.
The aim is to maximise acceptability amongst the most probable branches. Since all arguments from \textit{recommend} are attacked, there is 0 acceptability for that action; one argument from \textit{ignore} is attacked with 0.7 probability meaning \textit{ignore} is selected with the maximum acceptability of 0.3.

\section{Discussion}
\label{discussion}
Our goal here is to extend the typical approach to machine ethics, which is the assessment of a single action from the perspective of a single ethical theory, often without any account of probability or uncertainty.  
We have formalised Hansson's hypothetical retrospection procedure, systematising moral assessments as comparisons between consequences. This forms richer judgements beyond the evaluation of utilities.
Furthermore, our moral assessments are comparisons between retrospective justifications of hypothetical consequences.
One might ask how this differs from directly analysing the properties of consequences?
 For machines, it gives a procedure for selecting actions and providing justifications. For humans, it offers a resonance that allows us to make clearer judgements~\cite{hansson2013ethics}.
 It also allows us, in the future, to build on existing work for evaluating actions from the perspective of individual ethical theories and combining those judgements into arguments.  Essentially our proposal extends, rather than replaces, existing mechanisms for evaluating actions against a single ethical theory.

The retrospective procedure formalised by the critical questions resembles real life discussion: a claim against an argument and a chance to refute.
Say someone takes action $a_2$ in preference to $a_1$ and a principle is broken.  Retrospective argumentation through the critical questions produces a dialogue similar to the following:
\begin{center}
    \begin{enumerate}
        \item You should have chosen $a_1$ because it didn't break this moral principle.

        \item No, because there is a greater probability of breaking some other principle with $a_1$. If I was given the decision again, I would make the same choice.
    \end{enumerate}
\end{center}

Real life discussion may not be so civil, but if facts were agreed upon, this is the logical dialogue. Resonance with real life has utility for agent transparency and explainability, important for ethical AI~\cite{balasubramaniam2022transparency} and stakeholder buy-in.

The implementation is theory-neutral, allowing multiple principles and theories to be considered at once, more analogous to human decision-making.
Implementational work remains, not least the integration into a planning system to generate branches, but also evaluation against a wider range of ethical theories (e.g. Virtue Ethics) to see how easily they answer the critical questions.
We also wish to develop the evaluation of action's consequences along branches, not just at the branches end -- for instance, if someone is made unhappy as a consequence of some action, but then we compensate them by the end of the branch, can we ignore that we caused them (albeit temporary) unhappiness?

Implementations of hypothetical retrospection could be integrated into more general agent reasoning either as modules on top of an existing autonomous system, possibly similar to Arkin's governor architecture~\cite{arkin2008governing}.  Cardoso et. al have, for instance, considered how such ethical governors might integrate with BDI agents~\cite{CardosoEMAS21}.  Alternatively hypothetical retrospection could be implemented as a general decision-making process in which, for instance, the extent to which an action enables an agent to achieve or maintain goals could be included together with the arguments based upon ethical theories.  Systems of this kind -- in which all reasoning is encompassed within the ethical reasoning system can be seen in, for instance, the GenEth System~\cite{8500162} where ``maintain readiness" is treated as an ethical duty or the HERA system~\cite{lindner2017hera} where in~\cite{Dennis_Bentzen_Lindner_Fisher_2021} the system defaults to Utilitarianism to decide among actions all of which are considered equally valid according to some ethical theory.

Our current implementation has a fairly simple approach to the integration of ethical theories. Some theories are directly incompatible, potentially leading to ``worst of both worlds'' solutions.
Additionally, the use of utility classes needs careful handling. When utilities are of a greater class, they are prioritised, no matter how remote their probabilities.
Extending the Coin-Apple scenario, suppose an agent is offered a free apple every day -- as opposed to some number of apples all at once, or suppose the chance of winning the Hawaii holiday is extremely low, or both.
The justification for sacrificing a lifetime supply of apples for a small chance of a holiday is considerably weaker than sacrificing one apple for a 50/50 chance of a holiday.  Expected utility clearly has a part to play, even if the calculation of such utilities is non-trivial.  The difficultly in estimating utilities, and the fact that utilities may depend upon unknown factors such as a person's financial situation, mean there is uncertainty in the evaluation of state utilities which our framework currently does not address.
 
 There will be some computational complexity in searching and representing actions' potential branches of future development. In Section \ref{sec:hypotheticalRetrospection}, we note Hansson's principles for optimising search but it remains to be seen if this can be practically implemented to keep planning tractable for common problems.

 Nevertheless we believe the hypothetical retrospection framework practically handles many of the issues in machine ethics -- particularly the handling of uncertainty and the lack of any real agreement on the best moral theory.
  
 \section*{Acknowledgements}
 We would like to thank the University of Manchester for funding and EPSRC, under project Computational Agent Responsibility (EP/W01081X/1). 
 \section*{Open Data Statement}
 This work is licensed under a Creative Commons Attribution 4.0 International License. The tools/examples shown in this paper and instructions on reproducibility are openly available on GitHub at:
 \\\url{https://github.com/sameysimon/HypotheticalRetrospectionMachine}

\bibliographystyle{splncs04} 
\bibliography{myBibliography}
\end{document}